\documentclass[11pt,a4paper]{article}
\usepackage[hyperref]{acl2021}
\usepackage{times}
\usepackage{latexsym}

\usepackage{microtype}

\aclfinalcopy 

\usepackage[utf8]{inputenc} 
\usepackage[T1]{fontenc}    
\usepackage{hyperref}       
\usepackage{url}            
\usepackage{booktabs}       
\usepackage{amsfonts}       
\usepackage{nicefrac}       
\usepackage{microtype}      
\usepackage{amsmath}
\usepackage{amsthm}
\usepackage{mathtools}
\usepackage{dsfont}
\usepackage{xcolor}
\usepackage{graphicx}
\usepackage[nameinlink]{cleveref}
\usepackage{multirow}
\usepackage{siunitx}
\usepackage{caption}
\usepackage{subcaption}
\usepackage[nobiblatex]{xurl}
\usepackage{arydshln}

\def\eg{e.g.}

\usepackage{tikz}
\newcommand{\Arrow}[1]{%
    \parbox{#1}{\tikz{\draw[<-](0,0)--(#1,0);}}
}

\DeclareMathOperator{\accuracy}{Accuracy}
\DeclareMathOperator{\rad}{RAD}

\newcommand{\ync}{Y/N \Arrow{.2cm}C}
\newcommand{\ynhm}{Y/N \Arrow{.2cm}HM}
\newcommand{\ynwk}{Y/N \Arrow{.2cm}WK}

\title{Are VQA Systems RAD? \\ Measuring Robustness to Augmented Data with Focused Interventions}

\author{%
    Daniel Rosenberg \and Itai Gat \and Amir Feder \and Roi Reichart \\
    Technion - Israel Institute of Technology \\
    daniel.rnberg@gmail.com | \{itaigat@ | feder@campus. | roiri@\}technion.ac.il
}

\begin{document}
\maketitle

\begin{abstract}
    Deep learning algorithms have shown promising results in visual question answering (VQA) tasks, but a more careful look reveals that they often do not understand the rich signal they are being fed with. To understand and better measure the generalization capabilities of VQA systems, we look at their robustness to counterfactually augmented data. Our proposed augmentations are designed to make a focused intervention on a specific property of the question such that the answer changes. Using these augmentations, we propose a new robustness measure, Robustness to Augmented Data (RAD), which measures the consistency of model predictions between original and augmented examples. Through extensive experimentation, we show that RAD, unlike classical accuracy measures, can quantify when state-of-the-art systems are not robust to counterfactuals. We find substantial failure cases which reveal that current VQA systems are still brittle. Finally, we connect between robustness and generalization, demonstrating the predictive power of RAD for performance on unseen augmentations.\footnote{Our code and data are available at: \url{https://danrosenberg.github.io/rad-measure/}}
\end{abstract}

\section{Introduction}

In the task of Visual Question Answering (VQA), given an image and a natural language question about the image, a system is required to answer the question accurately \citep{vqa-dataset}. While the accuracy of these systems appears to be constantly improving \citep{fukui2016multimodal, yang2016stacked, lu2016hierarchical}, they are sensitive to small perturbations in their input and seem overfitted to their training data \citep{kafle2019challenges}. 

To address the problem of overfitting, the VQA-CP dataset was proposed \citep{vqa-cp-dataset}. It is a reshuffling of the original VQA dataset, such that the distribution of answers per question type (\eg, ``what color'', ``how many'') differs between the train and test sets. Using VQA-CP, \citet{kafle2019challenges} demonstrated the poor out-of-distribution generalization of many VQA systems. 
While many models were subsequently designed to deal with the VQA-CP dataset \citep{vqa-cp-rubi, vqa-cp-ensemble, vqa-cp-css, gat2020removing}, aiming to solve the out-of-distribution generalization problem in VQA, they were later demonstrated to overfit the unique properties of this dataset \citep{teney2020ood}. Moreover, no measures for robustness to distribution shifts have been proposed.

\begin{figure}[t]
    \centering
    \includegraphics[width=1\linewidth]{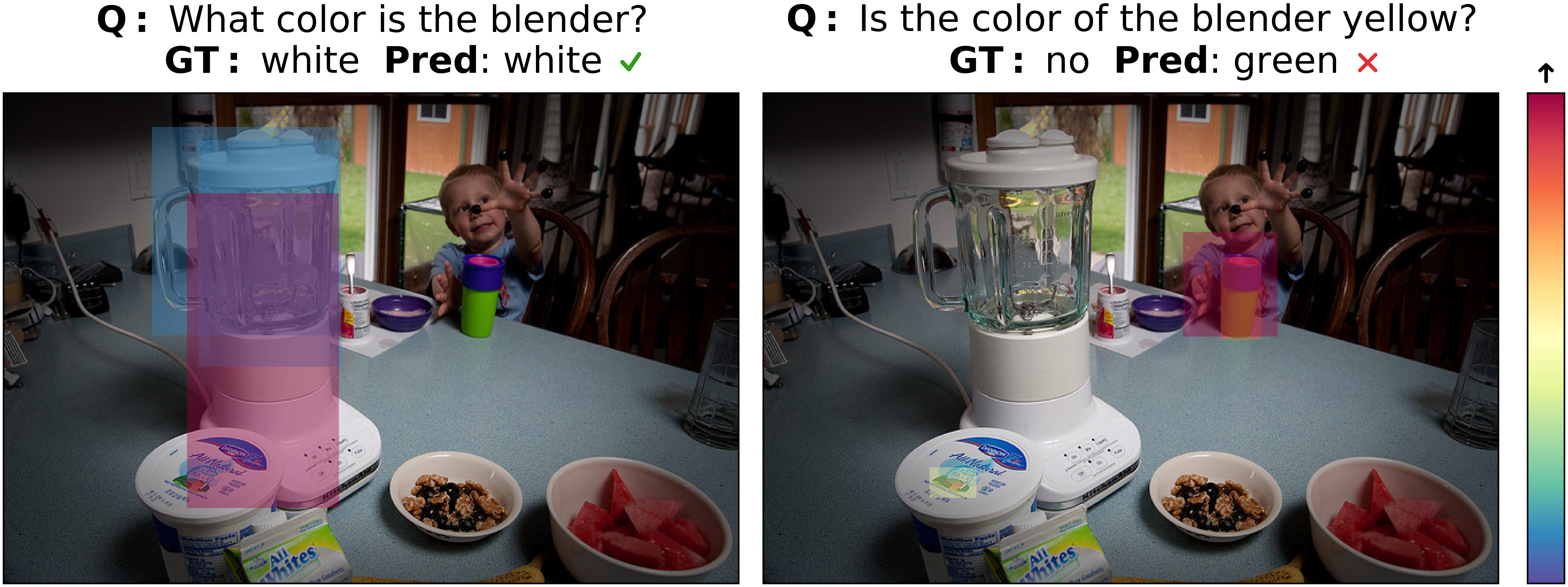}
    \caption{Predictions and attention maps of a state-of-the-art VQA-CP model over a VQA example (left) and its augmentation (right). A robust model should use the information it utilizes in the original example to correctly answer the augmented one.}
    \label{fig:teaser}
    \vspace{-0.3cm}
\end{figure}

In this work we propose a consistency-based measure that can indicate on the robustness of VQA models to distribution shifts. Our robustness measure is based on counterfactual data augmentations (CADs), which were shown useful for both training \citep{kaushik2019learning-the-difference} and evaluation \citep{garg2019counterfactual, agarwal2020towards}. CADs are aimed at manipulating a specific property while preserving all other information, allowing us to evaluate the robustness of the model to changes to this property.

For example, consider transforming a ``what color'' question to a ``yes/no'' question, as depicted in \Cref{fig:teaser}. The counterfactual reasoning for such a transformation is: ``what would be the question if it had a yes/no answer{?}''. While VQA models have seen many of both question types, their combination (yes/no questions about color) has been scarcely seen. If a model errs on such a combination, this suggests that to answer the original question correctly, the model uses a spurious signal such as the correlation between the appearance of the word ``color'' in the question and a particular color in the answer (e.g. here, color $\Rightarrow$ white). Further, this example shows that some models cannot even identify that they are being asked a ``yes/no'' question, distracted by the word ``color'' in the augmented question and answering ``green''.

Our robustness measure is named RAD: Robustness to (counterfactually) Augmented Data (\Cref{sec:rad_def}). RAD receives (image, question, answer) triplets, each augmented with a triplet where the question and answer were manipulated. It measures the consistency of model predictions when changing a triplet to its augmentation, i.e., the robustness of the model to (counterfactual) augmentations. We show that using RAD with focused interventions may uncover substantial weaknesses to specific phenomenon (\Cref{sec:robust_results}), namely, users are encouraged to precisely define their interventions such that they create \textit{counterfactual} augmentations. As a result, pairing RAD values with accuracy gives a better description of model behavior.

In general, to effectively choose a model in complex  tasks,  complementary  measures  are  required \cite{d2020underspecification}. Thus, it is important to have interpretable measures that are widely applicable. Note that in this work we manipulate only textual inputs - questions and answers, but RAD can be applied to any dataset for which augmentations are available. In particular, exploring visual augmentations would be beneficial for the analysis of VQA systems. Further, representation-level counterfactual augmentations are also valid, which is useful when generating meaningful counterfactual text is difficult \cite{feder2020causalm}.

Our augmentations (CADs) are generated semi-automatically (\Cref{sec:generation}), allowing us to directly intervene on a property of choice through simple templates. As in the above example, our augmentations are based on compositions of two frequent properties in the data (e.g., ``what color'' and ``yes/no'' questions), while their combination is scarce. Intuitively, we would expect a model with good generalization capacities to properly handle such augmentations. While this approach can promise coverage of only a subset of the examples in the VQA and VQA-CP datasets, it allows us to control the sources of the model's prediction errors. 

We conduct extensive experiments and report three key findings. First, for three datasets, VQA, VQA-CP, and VisDial~\citep{visdial}, models with seemingly similar accuracy are very different in terms of robustness, when considering RAD with our CADs (\Cref{sec:robust}). Second, we show that RAD with alternative augmentation methods, which prioritize coverage over focused intervention, cannot reveal the robustness differences. Finally, we show that measuring robustness using RAD with our CADs predicts the accuracy of VQA models on unseen augmentations, establishing the connection between robustness to our controlled augmentations and generalization (\Cref{sec:predictive_power}).

\section{Robustness to Counterfactuals}

In this section, we first present RAD (\Cref{sec:rad_def}), which measures model consistency on question-answer pairs and their augmented modifications. Then, we describe our template-based CAD generation approach (\Cref{sec:generation}), designed to provide control over the augmentation process.

\subsection{Model Robustness}\label{sec:rad_def}

We denote a VQA dataset with $\mathcal{U} = \left\{ (x_v,x_q,y) \in \mathcal{V} \times \mathcal{Q}  \times \mathcal{Y} \right\}$, where $x_v$ is an image, $x_q$ is a question and $y$ is an answer. We consider a subset $\mathcal{D} \subseteq \mathcal{U}$  for which we can generate augmentations. For an example $(x_v, x_q, y) \in \mathcal{D}$, we denote an augmented example as $(x_{v}, x_{q}', y') \in \mathcal{D}'$. In this paper we generate a single augmentation for each example in $\mathcal{D}$, resulting in a one-to-one correspondence between $\mathcal{D}$ and the dataset of modified examples $\mathcal{D}'$. We further define $J( \mathcal{D}; f)$ as the set of example indices for which a model $f$ correctly predicts $y$ given $x_v$ and $x_q$.

RAD assesses the proportion of correctly answered modified questions, among correctly answered original questions, and is defined as,
\begin{equation}
    \rad(\mathcal{D}, \mathcal{D}'; f) = \frac{|J(\mathcal{D}; f) \cap J(\mathcal{D}'; f)|}{|J( \mathcal{D}; f)|}.
\end{equation}
Note that RAD is in $[0, 1]$ and the higher the RAD of $f$ is, the more robust $f$ is. 

As original examples and their augmentations may differ in terms of their difficulty to the model, it is important to maintain symmetry between $\mathcal{D}$ and $\mathcal{D}'$. We hence also consider the backward view of RAD, defined as $\rad(\mathcal{D}', \mathcal{D}; f)$. For example, ``yes/no'' VQA questions are easier to answer compared to ``what color'' questions, as the former have two possible answers while the latter have as many as eight. Indeed, state-of-the-art VQA models are much more accurate on yes/no questions compared to other question types \citep{yu2019mcan}. Hence, if ``what color'' questions are augmented with ``yes/no'' counterfactuals, we would not expect $\rad(\mathcal{D}', \mathcal{D}; f) = 1$ as generalizing from ``yes/no'' questions ($\mathcal{D}'$) to ``what color'' questions ($\mathcal{D}$) requires additional reasoning capabilities.

RAD is not dependant on the accuracy of the model on the test set. A model may perform poorly overall but be very consistent on questions that it has answered correctly. Conversely, a model that demonstrates seemingly high performance may be achieving this by exploiting many dataset biases and be very inconsistent on similar questions.

\vspace{-0.1cm}
\subsection{Counterfactual Augmentations}\label{sec:generation}

In the VQA dataset there are three answer types: ``yes/no'', ``number'' (\eg, `2', `0') and ``other'' (\eg, `red`, `tennis'), and 65 question types (\eg, ``what color'', ``how many'', ``what sport''). In our augmentations, we generate ``yes/no'' questions from ``number'' and ``other'' questions.

For example, consider the question-answer pair ``What color is the vehicle? Red'', this question-answer pair can be easily transformed into ``Is the color of the vehicle red? Yes''. In general, ``what color'' questions can be represented by the template: ``What color is the \textit{<Subj>}? \textit{<Color>}''. To generate a new question, we first identify the subject (\textit{<Subj>}) for every ``what color'' question, and then integrate it into the template ``Is the color of the \textit{<Subj>} \textit{<Color>}? Yes''.  As the model was exposed to both ``what color'' and ``yes/no`` questions, we expect it to correctly answer the augmented question given that it correctly answers the original. Yet, this augmentation requires some generalization capacity because the VQA dataset contains very few yes/no questions about color.

Our templates are presented in \Cref{tab:templates} (see \Cref{tab:templates-examples} in the appendix for some realizations). The augmentations are counterfactual since we intervene on the question type, a prior that many VQA systems exploit~\citep{making-v-in-vqa-matter}, keeping everything else equal. The generation process is semi-automatic, as we had to first manually specify templates that would yield augmented questions that we can expect the model to answer correctly given that it succeeds on the original question. 

To achieve this goal, we apply two criteria: \textbf{(a)} The template should generate a grammatical English question; and \textbf{(b)} The generated question type should be included in the dataset, but not in questions that address the same semantic property as the original question. Indeed, yes/no questions are frequent in the VQA datasets, but few of them address color (first template), number of objects (second template), and object types (third template). When both criteria are fulfilled, it is reasonable to expect the model to generalize from its training set to the new question type. 

Criterion (a) led us to focus on yes/no questions since other transformations required manual verification for output grammaticality. While we could have employed augmentation templates from additional question types into yes/no questions, we believe that our three templates are sufficient for evaluating model robustness. Overall, our templates cover 11\% of the VQA examples (\Cref{sec:exp}).

\begin{table}
    \centering
    \small
    \begin{tabular}{lp{0.29\linewidth}p{0.36\linewidth}}
        \toprule
        {} &  Original & Augmented \\
        
        \midrule
        Y/N \Arrow{.2cm} C & What color is the \textit{<S>}? \textit{<C1>} & Is the color of the \textit{<S>} \textit{<C2>}? Yes/No \\

        \midrule
        Y/N \Arrow{.2cm}HM & How many \textit{<S>}? \textit{<N1>} & Are there \textit{<N2>} \textit{<S>}? Yes/No \\
        
        \midrule
        Y/N \Arrow{.2cm}WK & What kind of \textit{<S>} is this? \textit{<O1>} & Is this \textit{<S>} \textit{<O2>}? Yes/No \\
        
        \bottomrule
    \end{tabular}
    \vspace{-0.1cm}
    \caption{Our proposed template-based augmentations.}
    \label{tab:templates}
    \vspace{-0.4cm}
\end{table}

\setlength{\tabcolsep}{5pt}
\begin{table*}[t]
    \small
    \centering
    \begin{tabular}{llSSSSSSSc}
        \toprule
        \multirow{2}{*}{Dataset} & \multirow{2}{*}{Model\textbackslash $\mathcal{D}'$} & \multicolumn{7}{c}{$\rad(\mathcal{D}, \mathcal{D}')$ (\%)} & \multirow{2}{*}{Acc.} \\ \cmidrule{3-9} {} & &  {\ync{}} & {\ynhm{}} & {\ynwk{}} & {BT} & {Reph} & {L-ConVQA} & {CS-ConVQA} \\ \midrule
        \multirow{3}{*}{VQA-CP} & RUBi & 64.92 & 57.15 & 62.59 & 85.57 & 77.73 & 78.02 & 65.93 & 46.66 \\
         & LMH  &  1.01 & 22.82 & 50.10 & 83.68 & 75.04 & 64.54 & 50.65 & 53.72\\
         & CSS  & 0.94  & 11.73 & 39.95 & 77.54 & 68.89 & 10.67& 38.64  & 58.47\\
        \midrule
        \multirow{4}{*}{VQA} & BUTD & 67.15	& 58.68&	78.59&	87.43&	79.28&	75.78&	70.19 & 63.09\\
         & BAN & 74.40	& 62.45 &	82.51 &	88.17 &	81.14 &	79.37 &	70.18 & 65.92 \\
         & Pythia & 65.00&	60.61&	81.60&	88.42&	82.86&	77.02&	69.45& 64.56 \\
         & VisualBERT  & 79.99	& 68.29	&85.98	&88.52	&84.09	&82.09	&71.75 & 65.62 \\ \midrule
         \multirow{2}{*}{VisDial} & FGA & 31.36 & 57.69 & {-} & {91.42} & {-} & {-} & {-} & 53.07 \\
         & VisDialBERT & 62.08 & 56.06 & {-} & {94.04} & {-} & {-} & {-} & 55.78 \\ \bottomrule
    \end{tabular}
    \caption{RAD over our proposed augmentations (\ync{}, \ynhm{}, \ynwk{}) and alternatives (BT, Reph, ConVQA). The rows correspond to state-of-the-art models on VQA-CP (top), VQA (middle) and Visual Dialog (bottom). Reph and ConVQA were not created for VisDial, and it does not have ``what kind'' questions. The last column corresponds to validation accuracy.
    }
    \label{tab:rad_results}
\end{table*}
\setlength{\tabcolsep}{6pt}

\setlength{\tabcolsep}{5pt}
\begin{table*}[t]
    \small
    \centering
    \begin{tabular}{llSSSSSSSc}
        \toprule
        \multirow{2}{*}{Dataset} & \multirow{2}{*}{Model\textbackslash $\mathcal{D}'$} & \multicolumn{7}{c}{$\accuracy(\mathcal{D})$ (\%)} \\ \cmidrule{3-9} {} & &  {\ync{}} & {\ynhm{}} & {\ynwk{}} & {BT} & {Reph} & {L-ConVQA} & {CS-ConVQA} \\ \midrule
        \multirow{3}{*}{VQA-CP} & RUBi & 65.85 & 17.35 & 44.14 & 45.80 & 46.51 & 72.14 & 66.67 \\
         & LMH  &  68.87 & 44.24 & 50.58 & 52.35 & 53.78 & 65.07 & 61.76 \\
         & CSS  & 72.87 & 63.16 & 51.83 & 56.37 & 58.81 & 49.84 & 56.12  \\
        \midrule
        \multirow{4}{*}{VQA} & BUTD & 79.44 & 54.43 & 63.49 & 60.37 & 62.23 & 75.05 & 62.42 \\
         & BAN & 80.72 & 62.37 & 66.48 & 63.02 & 64.81 & 74.94 & 65.01 \\
         & Pythia & 81.62 & 57.49 & 64.42 & 61.69 & 63.88 & 74.55 & 63.79 & \\
         & VisualBERT  & 80.85 & 58.89 & 64.46 & 62.71 & 64.96 & 76.50 & 66.01 \\ \midrule
         \multirow{2}{*}{VisDial} & FGA & 55.62 & 40.00 & {-} & {61.53} & {-} & {-} & {-} \\
         & VisDialBERT & 68.99 & 50.77 & {-} & {63.47} & {-} & {-} & {-} \\ \bottomrule
    \end{tabular}
    \caption{Original accuracy over our proposed augmentations (\ync{}, \ynhm{}, \ynwk{}) and alternatives (BT, Reph, ConVQA). The rows correspond to state-of-the-art models on VQA-CP (top), VQA (middle) and Visual Dialog (bottom). Reph and ConVQA were not created for VisDial, and it does not have ``what kind'' questions.
    }
    \label{tab:accuracy_results}
    \vspace{-0.2cm}
\end{table*}
\setlength{\tabcolsep}{6pt}

\vspace{-0.05cm}
\section{Robustness with RAD and CADs}\label{sec:robust}

In the following, we perform experiments to test the robustness of VQA models to augmentations. We describe the experimental setup, and evaluate VQAv2, VQA-CPv2, VisDial models, each on our augmentations and on other alternatives.\footnote{The URLs of the software and datasets, and the implementation details are all provided in \Cref{sec:urls,,sec:model_settings}.}

\subsection{Experimental Setup} \label{sec:exp}

\paragraph{Baseline Augmentations} 
We compare our augmentations to three alternatives: VQA-Rephrasings (Reph, \citealp{vqa-rephrasings}), ConVQA \citep{con_vqa}, and back-translation (BT, \citealp{sennrich2016back-translation}). VQA-Rephrasings is a manual generation method, where annotators augment each validation question with three re-phrasings. ConVQA is divided into the L-ConVQA and CS-ConVQA subsets.
In both subsets, original validation examples are augmented to create new question-answer pairs.
L-ConVQA is automatically generated based on scene graphs attached to each image, and CS-ConVQA is manually generated by annotators. Finally, back-translation, translating to another language and back, is a high-coverage although low-quality approach to text augmentation. It was used during training and shown to improve NLP models \citep{sennrich2016back-translation}, but was not considered in VQA. We use English-German translations.

\paragraph{Models} 
The VQA-CP models we consider are RUBi~\citep{vqa-cp-rubi}, LMH~\citep{vqa-cp-ensemble} and CSS~\citep{vqa-cp-css}. The VQA models we consider are BUTD~\citep{butd-atten}, BAN~\citep{ban-vqa}, Pythia~\citep{pythia} and VisualBERT~\citep{visualbert}. For VisDial we use FGA~\citep{fga} and VisDialBERT~\citep{visdial-bert}. We trained all the  models using their official implementations.

\subsection{Results} \label{sec:robust_results}

\Cref{tab:rad_results} presents our main results. RAD values for all of our augmentations are substantially lower than those of the alternatives, supporting the value of our focused intervention approach for measuring robustness. The high RAD values for BT and Reph might indicate that VQA models are indeed robust to linguistic variation, as long as the answer does not change. Interestingly, our augmentations also reveal that VQA-CP models are less robust than VQA models. This suggests that despite the attempt to design more robust models, VQA-CP models still overfit their training data.

In VQA-CP, RUBi has the lowest accuracy performance in terms of its validation accuracy, even though it is more robust to augmentations compared with LMH and CSS. For VQA models, in contrast, BUTD has the lowest RAD scores on our augmentations and the lowest accuracy. VisualBERT, the only model that utilizes contextual word embeddings, demonstrates the highest robustness among the VQA models. 

Finally, while both VisDial models have similar accuracy, they have significantly different RAD scores on our augmentations. Specifically, VisDialBERT performs better than FGA on \ync{} augmentations. This is another indication of the value of our approach as it can help distinguish between two seemingly very similar models.

Complementary to the RAD values in \Cref{tab:rad_results} we also provide accuracies on original questions in \Cref{tab:accuracy_results}. Note that across all the original questions, except ConVQA questions, RUBi has the lowest accuracy while CSS has the highest accuracy. This trend is reversed when looking at RAD scores - CSS has the lowest score while RUBi has the highest score. This emphasizes the importance of RAD as a complementary metric, since considering only accuracy in this case would be misleading. Namely, RAD provides additional critical information for model selection.

\section{Measuring Generalization with RAD} \label{sec:predictive_power}

To establish the connection between RAD and generalization, we design experiments to demonstrate RAD's added value in predicting model accuracy on unseen modified examples. Concretely, we generate $45$ BUTD (VQA) and LMH (VQA-CP) instances, differing by the distribution of question types observed during training (for each model instance we drop between $10\%$ and $99\%$ of each of the question types ``what color'', ``how many'' and ``what kind'' from its training data; see \Cref{sec:regression_experiments} for exact implementation details). For each of the above models we calculate RAD values and accuracies in the following manner.

We split the validation set into two parts: $\mathcal{D}$ (features) and $\mathcal{T}$ (target). Consider a pool of four original question sets that are taken from their corresponding modifications: \ync{}, \ynhm{}, \ynwk{}, Reph. Then we have four possible configurations in which $\mathcal{D}$ is three sets from the pool and $\mathcal{T}$ is the remaining set.
For each model and for each configuration, we compute model accuracy on $\mathcal{D}$ ($\accuracy(\mathcal{D})$) and on the modifications of questions in $\mathcal{T}$ (the predicted variable $y(\mathcal{T}) = \accuracy(\mathcal{T}')$) which are modified with the target augmentation of the experiment. We also compute the RAD values of the model on the modified questions in $\mathcal{D}$ which are generated using the other three augmentations ($\rad(\mathcal{D}, \mathcal{D}'),$ and $\rad(\mathcal{D}', \mathcal{D})$). Then, we train a linear regression model using $\accuracy(\mathcal{D})$, $\rad(\mathcal{D}, \mathcal{D}'),$ and $\rad(\mathcal{D}', \mathcal{D})$, trying to predict $y(\mathcal{T})$. We perform this experiment four times, each using a different configuration (different augmentation type as our target), and average across the configurations.

\begin{table}[t]
    \small
    \centering
    \begin{tabular}{lccc}
        \toprule
        \multirow{2}{*}{Features\textbackslash Model} & \multicolumn{2}{c}{$R^2$} \\
        \cmidrule{2-3}
        {} & {LMH} \\
        \midrule
        $\accuracy(\mathcal{D}),$ & \multirow{3}{*}{$0.917 \pm 0.117$} \\
        $\rad(\mathcal{D}, \mathcal{D}'),$ \\
        $\rad(\mathcal{D}', \mathcal{D})$  \\
        \midrule
        $\accuracy(\mathcal{D})$ & $0.829 \pm 0.237$ \\
        \midrule
        $\rad(\mathcal{D}, \mathcal{D}')$ &  $0.899 \pm 0.133$ \\
        \midrule
        $\rad(\mathcal{D}', \mathcal{D})$ & $0.849 \pm 0.213$ \\
        \bottomrule
    \end{tabular}
    \caption{Linear regression experiments, predicting accuracy performance on unseen augmentation types.}
    \label{tab:regression}
    \vspace{-0.3cm}
\end{table}

\paragraph{Results}
\Cref{tab:regression} presents the average $R^2$ values and standard deviations over the four experiments. RAD improves the $R^2$ when used alongside the validation accuracy. Interestingly, a model's accuracy on one set of augmentations does not always generalize to other, unseen augmentations. Only when adding RAD to the regression model are we able to identify a robust model. Notably, for LMH the usefulness of RAD is significant, as it improves the $R^2$ by $11\%$. It also predicts performance better than validation accuracy when used without it in the regression. The standard deviations further confirm that the above claims hold over all configurations. These observations hold when running the same experiment with respect to the BUTD model, however, the improvements are smaller since the regression task is much easier with respect to this model ($R^2$ of 0.995 with all features).

\section{Conclusion}

We proposed RAD, a new measure that penalizes models for inconsistent predictions over data augmentations. We used it to show that state-of-the-art VQA models fail on CADs that we would expect them to properly address. Moreover, we have demonstrated the value of our CADs by showing that alternative augmentation methods cannot identify robustness differences as effectively. Finally, we have shown that RAD is predictive of generalization to unseen augmentation types.

We believe that the RAD measure brings substantial value to model evaluation and consequently to model selection. It encourages the good practice of testing on augmented data, which was shown to uncover considerable model weaknesses in NLP \cite{checklist}. Further, given visual augmentations, which we plan to explore in future work, or linguistic augmentations, RAD is applicable to any classification task, providing researchers with meaningful indications of robustness.

\section*{Acknowledgement}

This work was supported by funding from the Israeli ministry of Science and Technology.

\bibliography{refs}
\bibliographystyle{acl_natbib}

\clearpage
\appendix

\section{Dataset Statistics}
Please see \Cref{tab:dataset_counts} for the number of examples in each dataset that we use (VQA, VQA-CP and VisDial). We also report the number of augmentations we produce for each of our three augmentation types (\ync{}, \ynhm{} and \ynwk{}), alongside previous augmentation approaches used in our experiments (BT, Reph, L-ConVQA and CS-ConVQA).

\begin{table*}[t]
    \small
    \centering
    \begin{tabular}{lcccccccc}
        \toprule
        \multirow{2}{*}{Dataset} & \multicolumn{7}{c}{Augmentation Count} & \multirow{2}{*}{\shortstack{\\[1pt] Validation\\ Count}} \\ \cmidrule{2-8} {} & {\ync{}} & {\ynhm{}} & {\ynwk{}} & {BT} & {Reph} & {L-ConVQA} & {CS-ConVQA} \\ \midrule
        {VQA-CP} & 12,910 & 13,437 & 1,346 & 149,329 & 39,936 & 127,924 & 423 & 219,928 \\
        {VQA} & 12,835 & 10,233 & 1,654 & 138,043 & 121,512 & 127,924 & 1,365 & 214,354 \\
        {VisDial} & 516 & 130 & {-} & {1,136} & {-} & {-} & {-} & 20,640 \\ \bottomrule
    \end{tabular}
    \caption{Number of examples in each of the datasets we use.}
    \label{tab:dataset_counts}
\end{table*}

\setlength{\tabcolsep}{5pt}
\begin{table*}[ht]
    \small
    \centering
    \begin{tabular}{ccc}
        \toprule
        Yes/No $\leftarrow$ Colors & Yes/No $\leftarrow$ How Many & Yes/No $\leftarrow$ What Kind\\
        \midrule
        \text{What color is the cat? White}   &   \text{How many athletes are on the field? 5}  & \text{What kind of food is this? Breakfast} \\
        \color{blue} \text{Is the color of the cat white? Yes}     &    \color{blue} \text{Are there five athletes on the field? Yes}    &   \color{blue} \text{Is this food breakfast? Yes}\\
        
        \midrule
        \text{What color is the court? Green}     &       \text{How many dogs are in the  picture? 3}   &  \text{What kind of event is this? Skiing}\\
        \color{blue} \text{Is the color of the court green? Yes}       &       \color{blue} \text{Are there two dogs in the picture? No}   &   \color{blue} \text{Is this a skiing event? Yes} \\
        
        \midrule
        \text{What color is the vase? Blue}       &       \text{How many giraffes are walking around? 2}   & \text{What kind of animal is this? Cow}\\
        \color{blue} \text{Is the color of the vase red? No}  &   \color{blue} \text{Are there four giraffes walking around? No} & \color{blue} \text{Is this animal an elephant? No} \\
        
        \midrule 
        \text{What color is the man's hat? Red} &  \text{How many cakes are on the table? 0} & {What kind of building is this? Church} \\
        \color{blue} {Is the color of the man's hat red? Yes} & \color{blue} \text{Is there one cake on the table? No} & \color{blue} \text{Is this building a church? Yes} \\
        
        \midrule 
        \text{What color is the sky? Blue} &  \text{How many dogs? 1} & {What kind of floor is this? Wood} \\
        \color{blue} {Is the color of the sky blue? Yes} & \color{blue} \text{Are there zero dogs? No} & \color{blue} \text{Is this a wood floor? Yes} \\
        
        \bottomrule
    \end{tabular}
    \caption{Some realizations of our templates (defined in \Cref{tab:templates}). The black text (top) is the original question-answer pair and the blue text (bottom) is the corresponding augmented question-answer pair.}
    \label{tab:templates-examples}
\end{table*}
\setlength{\tabcolsep}{6pt}

\section{Our Augmentations} \label{sec:augmentations}

We describe the manual steps required to meet the desired standard for each augmentation type. For \ync{}, we filter out questions that start with ``What color is the''. For \ynhm{}, we use questions that starts with ``How many''. For \ynwk{}, we consider questions that match the pattern ``What kind of \textit{<S>} is this? \textit{<O1>}''. \autoref{tab:templates-examples} presents several realizations of the templates we define (see \Cref{sec:generation} for a discussion of these templates).

In \ynhm{}, we ensure that when the answer is `1', we use ``Is there ...'' instead of ``Are there ...''. We also ensure that the subsequent word to ``How many'' is a noun. We verify it is a noun using the part-of-speech tagger available through the spaCy library \citep{spacy}.

We allow the generation of both `yes' and `no' answers. Creating a modified question that is answered with a `yes' requires a simple permutation of words in the original question-answer pair, e.g., for \ync{}, take ``\textit{<C1>}'' = ``\textit{<C2>}'' (see \Cref{tab:templates}). Similarly, to generate a question that should be answered with a `no', we repeat the above process and change ``\textit{<C2>}''. In this case, we randomly pick an answer and replace it with the original answer with probability weighted with respect to the frequency in the data, among the pool of possible answers for the given augmentation type. When generating a new question, we first randomly decide whether to generate a `yes' or `no' question (with a probability of $0.5$ for each). Then, for example, if we choose to generate a `no', and ``\textit{<C1>}'' = ``red'', we have a $63\%$ chance of having ``\textit{<C2>}'' = ``blue''.

\section{URLs of Data and Code} \label{sec:urls}

\paragraph{Data} We consider three VQA datasets:
\begin{itemize}
    \setlength\itemsep{0.01em}
    \item The VQAv2 dataset \citep{making-v-in-vqa-matter}: \url{https://visualqa.org/}.
    \item The VQA-CPv2 dataset \citep{vqa-cp-dataset}: \url{https://www.cc.gatech.edu/grads/a/aagrawal307/vqa-cp/}.
    \item The VisDial dataset \citep{visdial}:
    \url{https://visualdialog.org/}
\end{itemize}

We also consider three previous augmentation methods:
\begin{itemize}
    \setlength\itemsep{0.01em}
    \item VQA-Rephrasings \citep{vqa-rephrasings}: \url{https://facebookresearch.github.io/VQA-Rephrasings/}.
    \item ConVQA \citep{con_vqa}: \url{https://arijitray1993.github.io/ConVQA/}.
    \item Back-translations \citep{sennrich2016back-translation}. We have generated these utilizing the transformers library \citep{wolf-etal-2020-transformers},
    \url{https://github.com/huggingface/transformers}.
    Specifically, we used two pre-trained translation models, English to German, and German to English:
    \url{https://huggingface.co/Helsinki-NLP/opus-mt-en-de}, \url{https://huggingface.co/Helsinki-NLP/opus-mt-de-en}.
\end{itemize}

\paragraph{Models} We consider nine models, where each model's code was taken from the official implementation. All implementations are via PyTorch \citep{pytorch}. 

The three VQA-CPv2 models:
\begin{itemize}
    \setlength\itemsep{0.01em}
    \item RUBi \citep{vqa-cp-rubi}: \url{https://github.com/cdancette/rubi.bootstrap.pytorch}.
    \item LMH \citep{vqa-cp-ensemble}: \url{https://github.com/chrisc36/bottom-up-attention-vqa}.
    \item CSS \cite{vqa-cp-css}: \url{https://github.com/yanxinzju/CSS-VQA}.
\end{itemize}

The four VQAv2 models:
\begin{itemize}
    \setlength\itemsep{0.01em}
    \item BUTD \citep{butd-atten}: \url{https://github.com/hengyuan-hu/bottom-up-attention-vqa}. \item BAN \citep{ban-vqa}: {\url{https://github.com/jnhwkim/ban-vqa}}.
    \item Pythia \citep{pythia}: Using the implementation in the MMF library \cite{singh2020mmf}, \url{https://github.com/facebookresearch/mmf}.
    \item VisualBERT \cite{visualbert}: Using the implementation in the MMF library.
\end{itemize}

And the two VisDial models:
\begin{itemize}
    \setlength\itemsep{0.01em}
    \item FGA \citep{fga}: \url{https://github.com/idansc/fga}.
    \item VisDialBERT \cite{visdial-bert}: \url{https://github.com/vmurahari3/visdial-bert}.
\end{itemize}

\section{Model Settings} \label{sec:model_settings}

We have trained the VQAv2 and the VQA-CPv2 models that we use, as pre-trained weights were not available for our requirements. For our evaluations, we require a model that is trained solely on the VQAv2 train set, such that we match the VQA-CPv2 settings, where there are only two sets, train and validation. In contrast, pre-trained models that are built for VQAv2 are trained on the VQAv2 training set and on the VQAv2 validation set together, as the dataset contains a third development set that is commonly used for validation.

We have trained six VQA models using the default hyper-parameters from their official implementations (URLs in \Cref{sec:urls}): RUBi, LMH, CSS, BUTD, BAN and Pythia. We trained the above models on a single Nvidia GeForce RTX 2080 Ti GPU, when the training time for each of the models was less than 12 hours. In addition, inference in this setting took less than an hour for all models.

The VisualBERT model is more computationally intensive, and we had to reduce the default batch size from 480 to 54 to fit it on our resources. Using three Nvidia GeForce RTX 2080 Ti GPUs for VisualBERT, training took 36 hours and inference took 4 hours.

For the VisDial models, FGA, and VisDialBERT, we have downloaded the pre-trained weights and used them solely for inference. On a single Nvidia GeForce RTX 2080 Ti GPU, inference took 15 minutes for FGA, and 8 hours for VisDialBERT. 

All the models we consider have less than 200M parameters.

When accuracies are reported on VQAv2 and on VQA-CP (\Cref{tab:accuracy_results,,tab:rad_results}) we use the VQA-accuracy metric \cite{vqa-dataset}. For VisDial we use the standard accuracy metric (denoted originally as R@1).

\section{Regression Experiments} \label{sec:regression_experiments}

We generate $45$ BUTD (VQA) instances and $45$ LMH (VQA-CP) instances. To generate different model instances, we create $45$ new training sets by removing examples from the original train set. For each of the three question types, ``what color'', ``how many'' and ``what kind'', we remove the following 15 percentage values of examples from the original train set: [10\%, 20\%, 30\%, 40\%, 50\%, 60\%, 70\%, 80\%, 90\%, 92\%, 95\%, 96\%, 97\%, 98\%, 99\%], resulting in $45$ new training sets. Then, each model instance is created by training on one of the $45$ training sets.

We split the validation set into two parts: $\mathcal{D}$ and $\mathcal{T}$. $\mathcal{D}$ is used to calculate the features in our linear regression model. We denote with $\mathcal{D}'_1$ the questions in $\mathcal{D}$ that can be modified using the \ync{} augmentation, after these questions were modified. Similarly, we define $\mathcal{D}'_2$, $\mathcal{D}'_3$, and $\mathcal{D}'_4$ for \ynhm{}, \ynwk{}, and Reph, respectively.

We average the $R^2$ of four linear regression experiments, when in each experiment we set a different $i$ ($i \in \{1,2,3,4\}$) for which $\mathcal{T} = \mathcal{D}'_i$ and use the remaining three templates to calculate our features. We denote the regression features with $x_1 = \accuracy(\mathcal{D})$, $x_2 = \rad(\mathcal{D}, \mathcal{D}') $, and $x_3 = \rad(\mathcal{D}', \mathcal{D})$, where $\rad(\mathcal{D}, \mathcal{D}')$ and $\rad(\mathcal{D}', \mathcal{D})$ are computed with respect to the three other templates ($j \in \{1,2,3,4\}, j \neq i$). The predicted label is $y(\mathcal{T}) = \accuracy(\mathcal{T})$.

Thus the equation for our regression is:
$$ y(\mathcal{T}) = b_1 x_1 + b_2 x_2 + b_3 x_3  + \epsilon \ . $$
We also perform three regression experiment for each feature alone:
$$ y(\mathcal{T}) = b x_k  + \epsilon,\quad k=1,2,3 \ ,$$
and compare the results of these experiments in \Cref{tab:regression}.

\end{document}